\documentclass[sigconf]{acmart}

\setcopyright{rightsretained}

\acmConference[ICVGIP2026]{17th Indian Conference on Computer Vision, Graphics and Image Processing}{December 21--25, 2026}{Kolkata, India}
\acmYear{2026}
\copyrightyear{2026}

\title{When Test-Time Adaptation Helps, Harms, or Becomes Inactive:
  A Condition-Level Study on CIFAR-10-C}

\author{Sreeja Guha Majumdar}
\affiliation{%
  \institution{Heritage Institute of Technology}
  \city{Kolkata}
  \state{West Bengal}
  \country{India}}
\email{sreejagm23@gmail.com}

\author{Aratrika Saha}
\affiliation{%
  \institution{Heritage Institute of Technology}
  \city{Kolkata}
  \state{West Bengal}
  \country{India}}
\email{aratrikasaha06@gmail.com}

\begin{document}
\begin{abstract}
Test-time adaptation (TTA) aims to improve model robustness under
distribution shift by adapting a source model using unlabeled test data.
Although methods such as TENT and EATA have demonstrated gains on corrupted
data, aggregate accuracy can obscure the conditions under which adaptation
fails or provides little benefit. We present a controlled comparison of three
TTA strategies---BatchNorm-statistics adaptation (BN-Adapt),
entropy-minimization adaptation (TENT), and reliability-filtered adaptation
(a scoped re-implementation of EATA)---against an unadapted source model on
the full CIFAR-10-C benchmark, covering 15 corruption types and 5 severity
levels (75 corruption--severity conditions, 10,000 images per condition).
Averaged across all conditions, TENT improves accuracy over the source model
by 13.29 percentage points (95\% bootstrap CI: [10.28, 16.29]), compared with
12.17 pp for BN-Adapt and 12.25 pp for EATA; all three improvements are highly
significant under a paired Wilcoxon signed-rank test ($p < 10^{-12}$).
However, the gains are not uniform: each method underperforms the source model
on 6--7 of the 75 conditions (8.0--9.3\%), with failures concentrated in
low-severity appearance corruptions such as brightness, fog, contrast, and
defocus blur, where the source model is already highly accurate. We further
find that EATA closely tracks the gradient-free BN-Adapt baseline, with a mean
absolute difference of 0.09 percentage points, compared with 1.08 percentage
points relative to TENT, indicating that reliability filtering can
substantially restrict effective adaptation and cause EATA to behave more like
a BatchNorm-statistics baseline than an entropy-minimization method. Beyond
the full benchmark, TENT exhibits non-monotonic accuracy with increasing test
batch size, while BN-Adapt and EATA improve monotonically across the tested
conditions; none of the three methods shows evidence of accuracy collapse in
the 256-batch continual stream without state reset. These results show that
aggregate accuracy alone can mask systematic TTA failure modes and motivate
condition-level evaluation of when adaptation helps, harms, or becomes
effectively inactive.
\end{abstract}

\ccsdesc[500]{Computing methodologies~Neural networks}
\ccsdesc[300]{Computing methodologies~Image processing}
\ccsdesc[300]{Computing methodologies~Computer vision}

\keywords{test-time adaptation, distribution shift, CIFAR-10-C, robustness,
  entropy minimization}

\maketitle

\section{Introduction}

Deep image classifiers trained on a fixed source distribution can suffer substantial
accuracy degradation when the test distribution shifts because of corruption,
sensor noise, or environmental changes. Test-time adaptation (TTA) addresses
this setting by adapting a pretrained model using unlabeled test data, without
requiring access to source data or test labels. TENT is a widely used fully
test-time adaptation method that minimizes the entropy of model predictions on
incoming test batches while adapting batch-normalization statistics and
channel-wise affine parameters \cite{wang2021tent}. Related approaches have
also investigated adaptation through batch-normalization statistics alone
\cite{nado2020,schneider2020}, while EATA introduces sample-selection
mechanisms and regularization to make entropy-based adaptation more efficient
and reduce undesirable adaptation \cite{niu2022eata}.

CIFAR-10-C provides a standardized benchmark for evaluating robustness to
common image corruptions, consisting of 15 corruption types evaluated at five
severity levels \cite{hendrycks2019}. Although aggregate accuracy over the
benchmark provides a useful measure for comparing methods, such aggregation can
obscure how adaptation behaves under individual corruption--severity
conditions. In particular, an overall improvement does not establish that
adaptation is beneficial for every corruption or severity, nor does it reveal
whether a method can reduce accuracy relative to the unadapted source model
under specific conditions.

In this work, we do not propose a new adaptation method. Instead, we perform a
controlled empirical comparison of three adaptation strategies---BN-Adapt,
TENT, and a scoped re-implementation of EATA---against an unadapted source
model on the full CIFAR-10-C benchmark. We evaluate all 15 corruption types
and five severity levels, yielding 75 corruption--severity conditions, and
examine performance both in aggregate and at the individual-condition level.
This allows us to identify conditions in which adaptation provides substantial
benefit as well as conditions in which it provides little benefit or reduces
accuracy relative to the source model.

Our evaluation focuses on four questions:
\begin{enumerate}
\item how much each adaptation method improves aggregate accuracy relative to
the unadapted source model, together with bootstrap confidence intervals and a
paired statistical significance test;
\item how these gains vary across corruption categories and severity levels,
including an explicit identification of conditions in which adaptation reduces
accuracy;
\item how each method responds to changes in test batch size; and
\item whether adaptation remains stable over a continual test stream when
model state is carried across batches without reset.
\end{enumerate}

Across the 75 corruption--severity conditions, all three methods substantially
improve mean accuracy over the unadapted source model, but the benefit is not
uniform. Each method also produces a small number of conditions in which
accuracy falls below the source model, with these failures concentrated among
low-severity appearance-related corruptions for which the source model is
already highly accurate. We further examine the relationship between the
methods and find that EATA closely tracks the gradient-free BN-Adapt baseline
in our experimental setting, suggesting that its reliability filtering can
substantially restrict the samples contributing to effective adaptation.

We additionally observe that TENT exhibits non-monotonic behavior as test
batch size increases, whereas BN-Adapt and EATA are more consistent across the
tested batch sizes. None of the three methods shows evidence of accuracy
collapse in our 256-batch continual adaptation stream without state reset.
Taken together, these results indicate that aggregate benchmark accuracy alone
can mask systematic condition-level differences in TTA behavior. Our findings
motivate evaluation protocols that explicitly examine when adaptation helps,
harms, or becomes effectively inactive, rather than relying solely on a single
aggregate accuracy measure.

\section{Related Work}

\subsection{Test-Time Adaptation}

Test-time adaptation studies the adaptation of pretrained models to a shifted
test distribution using unlabeled test data. TENT performs fully test-time
adaptation by minimizing the entropy of predictions on incoming batches and
updating a restricted set of model parameters associated with
batch normalization \cite{wang2021tent}. This provides a lightweight approach
to adapting a source model without requiring target-domain labels.

An alternative line of work considers adaptation through batch-normalization
statistics. Nado et al.~evaluate prediction-time batch normalization under
covariate shift \cite{nado2020}, while Schneider et al.~study covariate-shift
adaptation as a means of improving robustness to common corruptions
\cite{schneider2020}. These approaches provide a useful gradient-free
reference point for evaluating entropy-based adaptation.

\subsection{Efficient Adaptation}

EATA extends entropy-based test-time adaptation by restricting adaptation to
reliable and non-redundant samples and incorporating an anti-forgetting
mechanism \cite{niu2022eata}. In contrast to TENT, which applies its
entropy-minimization objective to the incoming adaptation batches, EATA
selectively determines which samples contribute to the update.

This distinction is particularly relevant to our analysis because a reliability
filter can alter not only the efficiency of adaptation but also its effective
behavior. We therefore compare EATA with both TENT and the gradient-free
BN-Adapt baseline to determine whether the filtered method exhibits behavior
closer to entropy-minimization adaptation or to batch-statistics adaptation
under the conditions considered in our experiments.

\subsection{Continual and Stability-Aware Adaptation}
\label{subsec:continual-related-work}

A related line of work targets adaptation over long, non-stationary test
streams rather than a single batch or a fixed corruption condition. CoTTA
addresses continually changing target domains by combining weight-averaged and
augmentation-averaged pseudo-labels with stochastic restoration of a randomly
selected subset of parameters to their source values at each step, explicitly
targeting error accumulation and catastrophic forgetting over long adaptation
horizons \cite{wang2022cotta}. SAR pursues a related stability goal, building
on EATA-style sample filtering with a sharpness-aware parameter update and a
model-recovery mechanism triggered when the batch loss collapses, so as to
prevent adaptation from diverging under small or class-imbalanced batches
\cite{niu2023sar}. Our long-stream protocol
(Section~\ref{subsec:additional-protocols}) is motivated by the same concern
that drives both methods, though we evaluate the standard TENT and EATA
updates rather than these stability-oriented variants, and find no evidence of
collapse at the stream length we test.

\subsection{Alternative Test-Time Objectives}

Not every test-time adaptation method relies on entropy minimization or
batch-normalization statistics. SHOT adapts a source-trained feature extractor
to an unlabeled target set using information-maximization and pseudo-labeling
objectives, in an offline, dataset-level setting rather than the online,
per-batch setting we study \cite{liang2020shot}. T3A instead leaves the
feature extractor and classifier weights unchanged entirely, adjusting only
class prototypes computed from unlabeled test features at inference time
\cite{iwasawa2021t3a}. These methods illustrate that the
entropy-minimization and batch-normalization-adaptation strategies compared in
this paper occupy only part of a considerably larger design space; recent
surveys provide a broader overview of that space
\cite{wang2023otta,xiao2024ttasurvey}.

\subsection{Corruption Robustness and Condition-Level Evaluation}

CIFAR-10-C was introduced as a benchmark for evaluating the robustness of
image classifiers against common corruptions and perturbations
\cite{hendrycks2019}. The benchmark contains 15 corruption types, each
evaluated at five severity levels, providing a structured test bed for
studying performance under distribution shift.

Our evaluation uses this structure to move beyond a single aggregate accuracy
value. We report performance separately for each corruption--severity
condition and explicitly identify cases in which adaptation improves or
reduces accuracy relative to the source model. We additionally examine
sensitivity to test batch size and stability under a continual adaptation
stream without state reset. This condition-level perspective complements
aggregate robustness evaluation by showing where the behavior of TTA methods
is consistent across shifts and where adaptation can become ineffective or
harmful.

\subsection{Source Model and CIFAR-10}

Our experiments use CIFAR-10 as the underlying image-classification dataset
\cite{krizhevsky2009} and evaluate robustness using its corrupted counterpart,
CIFAR-10-C \cite{hendrycks2019}. The source classifier is a Wide Residual
Network, following the architecture family introduced by Zagoruyko and
Komodakis \cite{zagoruyko2016}.

\section{Method}

\subsection{Problem Formulation}

We consider fully test-time adaptation of a pretrained image classifier under
distribution shift. Let $f_{\theta}$ denote a classifier with parameters
$\theta$, and let $(x_i,y_i)$ denote a target-domain test sample, where
$y_i$ is available only for evaluation and is not used during adaptation. Given
an incoming unlabeled test batch
$\mathcal{B}_t=\{x_1,\ldots,x_B\}$, a test-time adaptation method updates a
subset of the model parameters using only the predictions
$f_{\theta}(x_i)$ and statistics computed from the current test batch.

The unadapted source model provides the reference performance:
\begin{equation}
  \hat{y}_i = \arg\max_{c} p_{\theta_0}(c \mid x_i),
  \label{eq:source-prediction}
\end{equation}
where $\theta_0$ denotes the parameters obtained during source-domain
training. For adaptation-based methods, the model state is updated
sequentially as test batches arrive:
\begin{equation}
  \theta_t =
  \mathcal{A}\left(\theta_{t-1},\mathcal{B}_t\right),
  \label{eq:adaptation-update}
\end{equation}
where $\mathcal{A}$ denotes the adaptation procedure. No target labels are
provided to $\mathcal{A}$.

Our study does not introduce a new adaptation objective or algorithm.
Instead, we instantiate three representative adaptation procedures---BN-adapt,
TENT, and EATA---and compare them against the fixed source model under a
controlled evaluation protocol. This formulation allows all methods to be
evaluated under the same target batches, model architecture, and corruption
conditions.

\subsection{Adaptation Procedures}

\paragraph{Source model.}
The source model is evaluated without any test-time adaptation. Its parameters
remain fixed throughout evaluation and therefore provide the reference against
which adaptation gains and failures are measured.

\paragraph{BN-adapt.}
BN-adapt re-estimates batch-normalization statistics from each incoming test
batch without performing gradient-based parameter updates. The resulting
normalization statistics are therefore adapted to the observed target batch,
while the learned network parameters remain unchanged
\cite{nado2020,schneider2020}.

\paragraph{TENT.}
TENT re-estimates batch-normalization statistics and updates only the
channel-wise affine parameters by minimizing the mean prediction entropy of
the current test batch \cite{wang2021tent}. Its batch objective is
\begin{equation}
  \mathcal{L}_{\mathrm{ent}}(\theta;\mathcal{B}_t)
  = -\frac{1}{B}\sum_{i=1}^{B}\sum_{c=1}^{C}
  p_{\theta}(c\mid x_i)\log p_{\theta}(c\mid x_i),
  \label{eq:tent-entropy}
\end{equation}
where $B$ is the test-batch size and $C$ is the number of classes. The update
is performed once for each incoming batch.

\paragraph{EATA.}
EATA extends entropy-minimization adaptation by filtering samples before they
contribute to the adaptation update and incorporating an anti-forgetting
regularizer applied to the entropy-minimization update \cite{niu2022eata}. In
our experiments, we use a scoped re-implementation of EATA that follows the
same overall adaptation setting as TENT while applying its reliability-based
sample selection and anti-forgetting mechanism. The purpose is to evaluate
whether this filtering produces behavior that is substantially different from
TENT or instead restricts effective adaptation toward the behavior of the
gradient-free BN-adapt baseline.

\section{Experiments}

\paragraph{Datasets and evaluation protocol.}
We evaluate on CIFAR-10-C \cite{hendrycks2019}, the standard
benchmark for corruption robustness, which applies 15 corruption types
(Gaussian noise, shot noise, impulse noise, defocus blur, glass blur, motion
blur, zoom blur, snow, frost, fog, brightness, contrast, elastic transform,
pixelate, and JPEG compression) at 5 severity levels to the CIFAR-10 test
set \cite{krizhevsky2009}. This gives a $15\times5=75$ (corruption, severity)
grid, with $N=10{,}000$ images evaluated per condition. For each condition,
the source model and all three adaptation methods (BN-adapt, TENT, EATA) are
evaluated on the identical sequence of test batches, with batch size
$B=200$; a fresh copy of the source model is instantiated per method per
condition, so there is no state leakage across methods or across conditions.
Beyond this main grid, we additionally run a batch-size sensitivity study and
a long-stream protocol, described in
Section~\ref{subsec:additional-protocols}.

\paragraph{Baselines and metrics.}
We compare four conditions: the fixed source model, BN-adapt, TENT, and our
EATA re-implementation. For each we report (i) mean classification accuracy
across the 75-condition grid; (ii) \emph{harm rate}, the percentage of
conditions on which an adaptation method's accuracy falls below the source
model's accuracy on that same condition; (iii) severity-stratified mean
$\Delta$accuracy (adapted $-$ source), aggregated into low (severity 1--2),
mid (severity 3), and high (severity 4--5) bins; (iv) worst-case accuracy,
the minimum accuracy attained by each method across all corruption types;
(v) a paired Wilcoxon signed-rank test comparing each method's per-condition
accuracy against the source model, paired by (corruption, severity); and
(vi) a bootstrap confidence interval on the mean $\Delta$accuracy. We
additionally report a corruption-category breakdown (noise, blur, weather,
digital, following the standard CIFAR-10-C taxonomy) of $\Delta$accuracy for
each method.

\paragraph{Implementation details.}
The source model is a WideResNet-28-10 \cite{zagoruyko2016}, obtained as the
``Standard'' pretrained CIFAR-10 checkpoint from the RobustBench model zoo
under the \texttt{corruptions} threat model \cite{croce2021robustbench}; it is
trained with standard (non-adversarial) supervision on clean CIFAR-10 and is
not fine-tuned further, constituting $\theta_0$ in
Eq.~\eqref{eq:source-prediction}. For TENT and EATA, only the channel-wise
batch-normalization affine parameters (scale and shift) are updated, using
Adam \cite{kingma2014adam} with learning rate $10^{-3}$, one gradient step
per incoming batch, and no episodic reset (adapted parameters persist across
consecutive batches within a condition). For all three adaptation methods,
BN layers use current-batch statistics rather than the source running
statistics. EATA additionally applies an entropy-based reliability filter
with threshold $e_{\text{margin}} = 0.4\ln C$ ($C=10$ classes,
$e_{\text{margin}} \approx 0.92$ nats), discarding samples whose predictive
entropy exceeds this threshold; a redundancy filter that discards samples
whose softmax output has cosine similarity above $d_{\text{margin}}=0.05$ to
a running average of previously-adapted samples' outputs; and an
anti-forgetting regularizer weighted by $\lambda_{\text{fisher}}=2000$,
anchored at the source BN parameters, with diagonal Fisher information
estimated from 20 batches (4{,}000 images, batch size 200) of clean,
uncorrupted CIFAR-10 test data.

\subsection{Additional Protocols}
\label{subsec:additional-protocols}

Beyond the main 75-condition grid, we run two further protocols on a
representative subset of six corruptions (Gaussian noise, motion blur, fog,
contrast, JPEG compression, and snow). First, a batch-size sensitivity study
repeats each of BN-adapt, TENT, and EATA at severities 3 and 5 with test batch
sizes of 32, 128, and 512. Second, a long-stream protocol concatenates 256
consecutive batches drawn from these six corruptions at severity 5 (each
corruption block is 32 batches, and the six-corruption cycle repeats) into a
single continuous adaptation run with no state reset between blocks, to test
whether accuracy degrades as adaptation continues indefinitely.

\section{Results}

\subsection{Overall Accuracy Across Severities}

\begin{figure}[t]
  \centering
  \includegraphics[width=\linewidth]{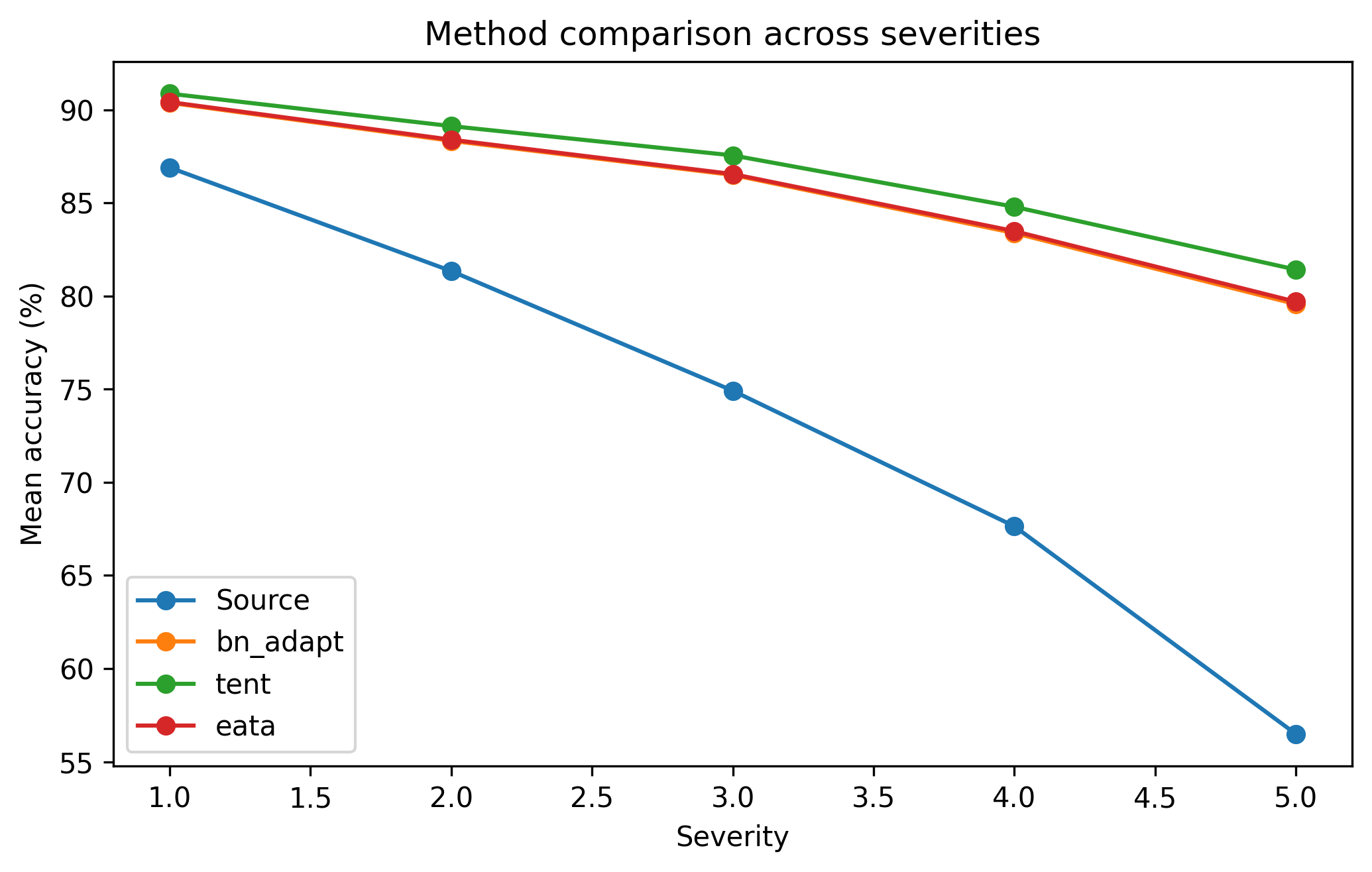}
  \caption{Mean accuracy versus corruption severity for the source model and
    each adaptation method. All three adaptation methods track closely and
    diverge from the source model as severity increases.}
  \Description{Line chart of mean accuracy against severity levels one through
    five for Source, BN-adapt, TENT, and EATA.}
  \label{fig:severity-accuracy}
\end{figure}

Table~\ref{tab:severity-accuracy} reports mean accuracy at each severity level,
averaged across all 15 corruption types. All three adaptation methods
substantially outperform the unadapted source model at every severity, and the
gap widens as corruption severity increases: at severity 1, TENT improves on
source by 3.97 percentage points (86.90\% to 90.87\%); at severity 5, the gap
grows to 24.93 points (56.49\% to 81.42\%). This trend, visualized in
Figure~\ref{fig:severity-accuracy}, replicates the qualitative pattern reported
in the original TENT and EATA papers: test-time adaptation is most valuable
exactly where the source model is weakest.

\begin{table}[t]
  \caption{Mean classification accuracy by corruption severity, averaged
    across all 15 corruption types (75 conditions total, 10,000 images per
    condition).}
  \label{tab:severity-accuracy}
  \centering
  \small
  \begin{tabular}{lrrrr}
    \toprule
    Severity & Source & BN-adapt & TENT & EATA \\
    \midrule
    1 & 86.90 & 90.37 & \textbf{90.87} & 90.41 \\
    2 & 81.34 & 88.33 & \textbf{89.12} & 88.39 \\
    3 & 74.92 & 86.50 & \textbf{87.55} & 86.55 \\
    4 & 67.64 & 83.38 & \textbf{84.79} & 83.48 \\
    5 & 56.49 & 79.56 & \textbf{81.42} & 79.69 \\
    \bottomrule
  \end{tabular}
\end{table}

\subsection{Statistical Robustness of the Gains}

To assess whether the average gains in Table~\ref{tab:severity-accuracy} are
reliable rather than an artifact of a few favorable conditions, we compute a
paired-bootstrap 95\% confidence interval on the mean accuracy delta over source
(10,000 resamples over the 75 corruption--severity conditions) and a paired
Wilcoxon signed-rank test comparing each method's per-condition accuracy against
the source model's. Table~\ref{tab:statistical-gains} reports both. All three
methods show a highly significant improvement over source ($p < 10^{-12}$ in
every case). TENT has the highest point estimate for mean gain, but its
confidence interval overlaps substantially with those of BN-adapt and EATA, so
the ranking among the three adaptation methods themselves should not be treated
as a confidently established fact; only their improvement over the unadapted
source model is unambiguous at this sample size.

\begin{table*}[t]
  \caption{Mean accuracy improvement over source, bootstrap 95\% confidence
    interval, and paired Wilcoxon signed-rank test $p$-value, computed over the
    75 corruption--severity conditions.}
  \label{tab:statistical-gains}
  \centering
  \begin{tabular}{lccc}
    \toprule
    Method & Mean $\Delta$ vs. Source (pp) & 95\% CI (bootstrap) & Wilcoxon $p$-value \\
    \midrule
    BN-adapt & 12.17 & [9.40, 15.10] & $5.3 \times 10^{-13}$ \\
    TENT & \textbf{13.29} & [10.28, 16.29] & $3.8 \times 10^{-13}$ \\
    EATA & 12.25 & [9.41, 15.21] & $4.9 \times 10^{-13}$ \\
    \bottomrule
  \end{tabular}
\end{table*}

\subsection{Where the Gains Concentrate: Category and Severity}

\begin{figure}[t]
  \centering
  \includegraphics[width=\linewidth]{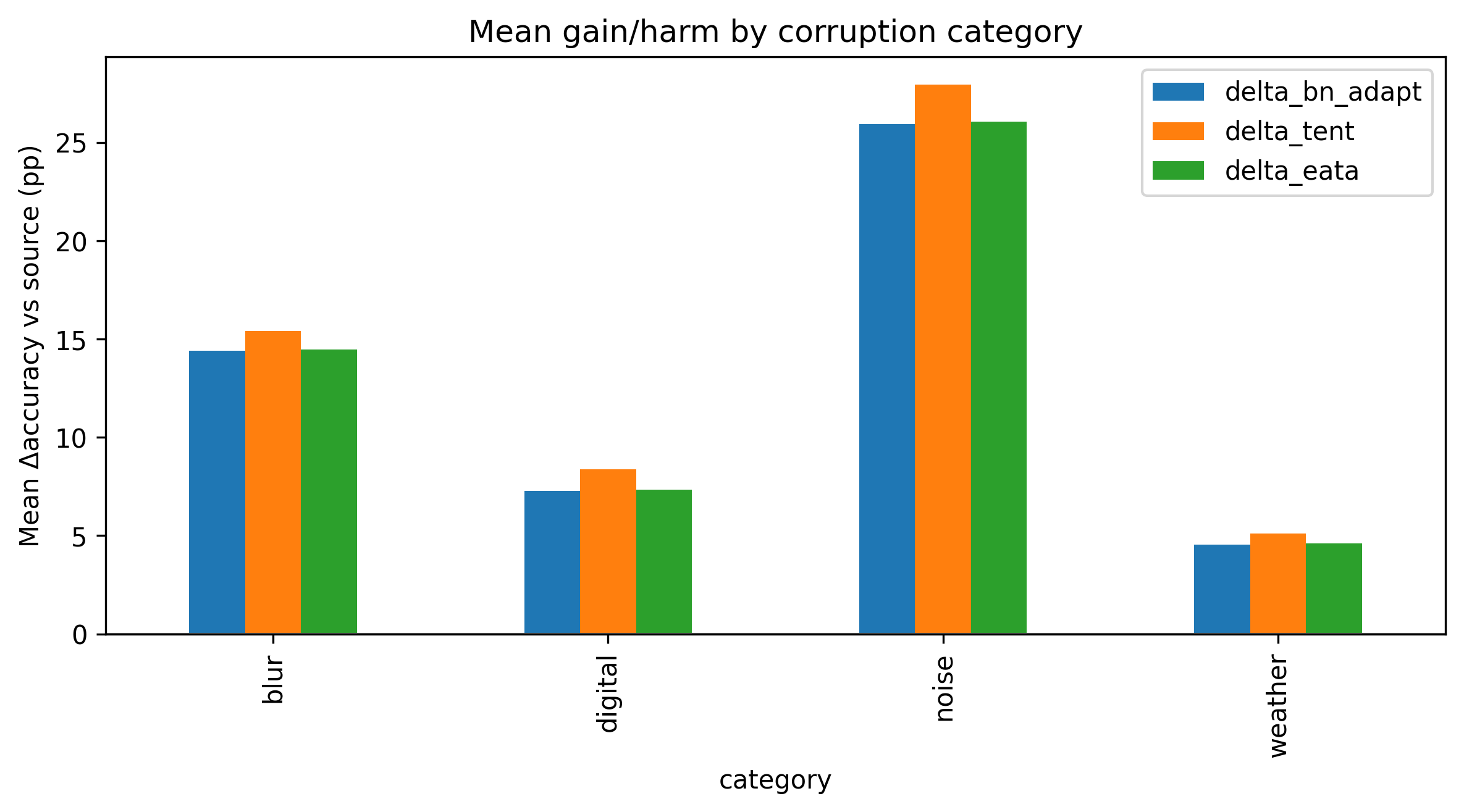}
  \caption{Mean accuracy gain over source by corruption category. Noise
    corruptions benefit most; weather corruptions benefit least and contain
    most conditions where adaptation underperforms source.}
  \Description{Grouped bar chart of accuracy gains for BN-adapt, TENT, and EATA
    across blur, digital, noise, and weather corruption categories.}
  \label{fig:category-gains}
\end{figure}

Breaking the mean gain down by corruption category
(Table~\ref{tab:category-gains} and Figure~\ref{fig:category-gains}) shows that
adaptation helps far more on noise corruptions (+26 to +28 pp) and blur
corruptions (+14 to +15 pp) than on digital (+7 to +8 pp) or weather
corruptions (+5 pp). This is consistent with the severity-stratified view in
Table~\ref{tab:severity-buckets}: gains are modest at low severity (+5 to +6 pp
for severities 1--2) and grow substantially at high severity (+19 to +21 pp for
severities 4--5). The two views are related: weather and digital corruptions in
this benchmark tend to leave the source model comparatively accurate even at
high severity, so there is simply less headroom for any adaptation method to
recover.

\begin{table}[t]
  \caption{Mean accuracy improvement over source by corruption category,
    averaged across constituent corruption types and all five severities.}
  \label{tab:category-gains}
  \centering
  \small
  \begin{tabular}{lrrr}
    \toprule
    Category & BN-adapt & TENT & EATA \\
    \midrule
    Noise & 25.94 & \textbf{27.95} & 26.06 \\
    Blur & 14.40 & \textbf{15.42} & 14.46 \\
    Digital & 7.26 & \textbf{8.37} & 7.33 \\
    Weather & 4.55 & \textbf{5.10} & 4.59 \\
    \bottomrule
  \end{tabular}
\end{table}

\begin{table}[t]
  \caption{Mean accuracy improvement over source by severity bucket, averaged
    across all 15 corruption types.}
  \label{tab:severity-buckets}
  \centering
  \small
  \begin{tabular}{lrrr}
    \toprule
    Severity bucket & BN-adapt & TENT & EATA \\
    \midrule
    Low (1--2) & 5.23 & \textbf{5.87} & 5.28 \\
    Mid (3) & 11.58 & \textbf{12.63} & 11.63 \\
    High (4--5) & 19.41 & \textbf{21.04} & 19.52 \\
    \bottomrule
  \end{tabular}
\end{table}

\subsection{Per-Condition Failure Analysis}

\begin{figure*}[t]
  \centering
  \begin{minipage}[t]{0.32\textwidth}
    \centering
    \includegraphics[width=\linewidth]{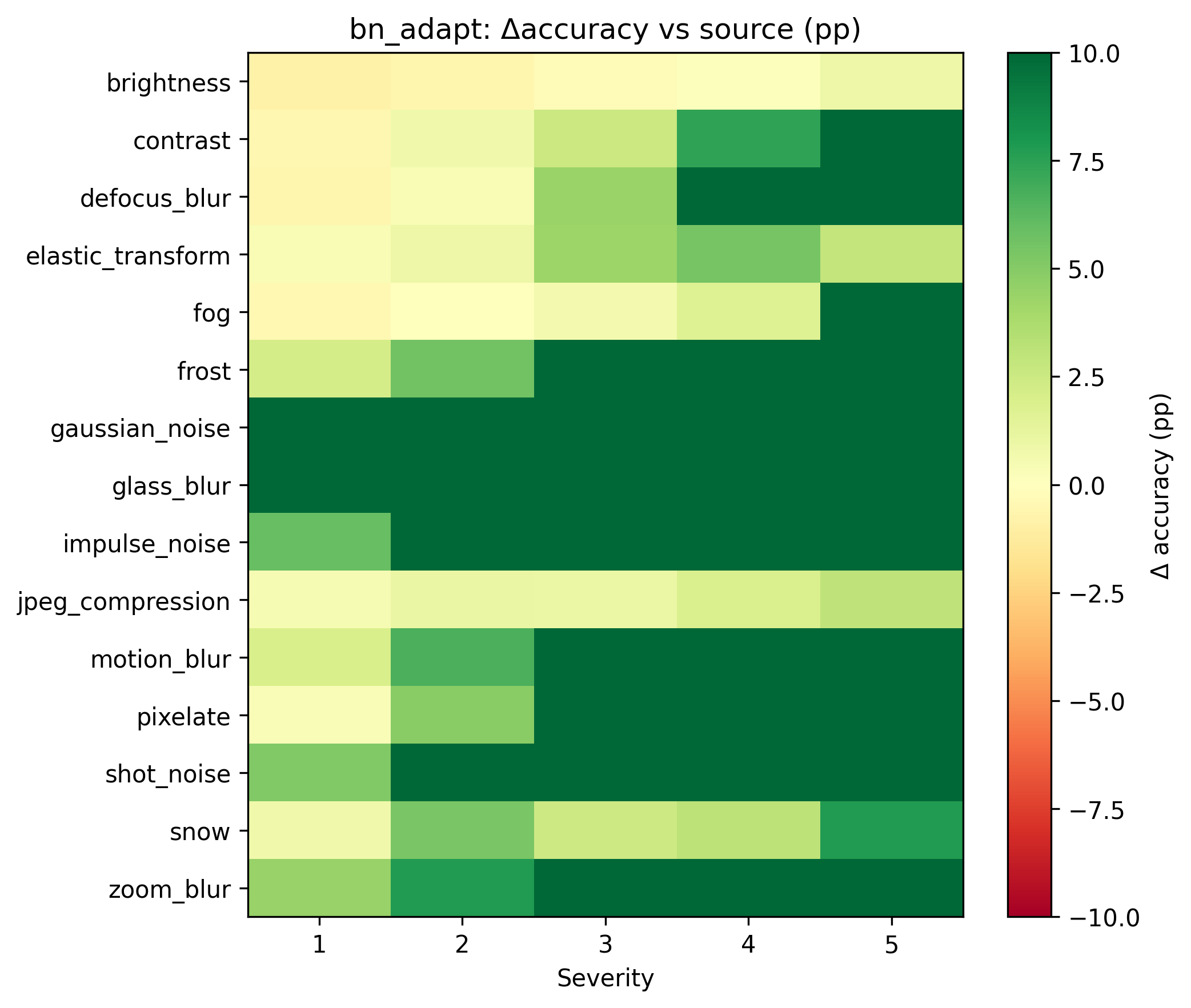}
  \end{minipage}\hfill
  \begin{minipage}[t]{0.32\textwidth}
    \centering
    \includegraphics[width=\linewidth]{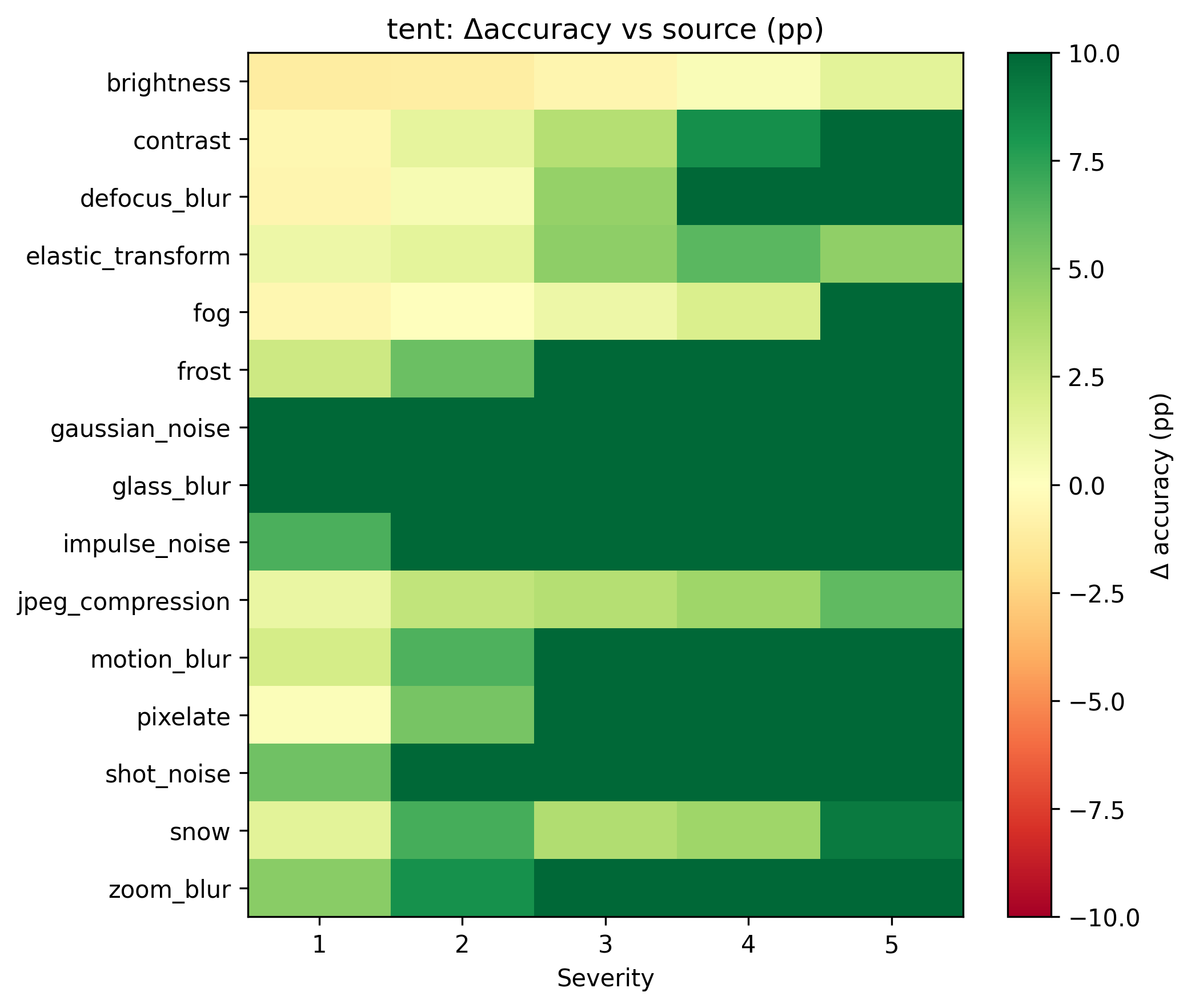}
  \end{minipage}\hfill
  \begin{minipage}[t]{0.32\textwidth}
    \centering
    \includegraphics[width=\linewidth]{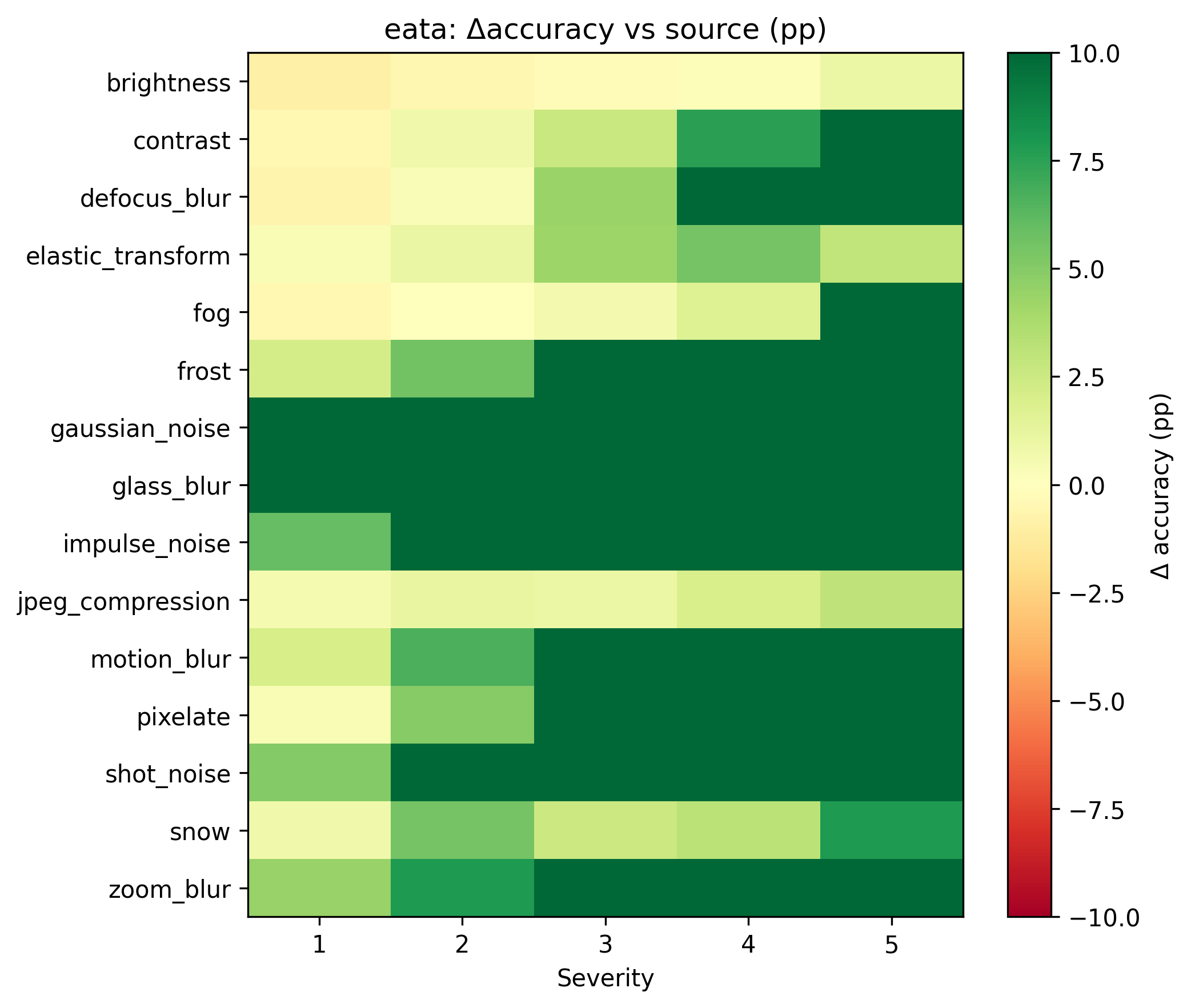}
  \end{minipage}
  \caption{Per-corruption, per-severity accuracy delta versus source
    (percentage points) for BN-adapt (left), TENT (center), and EATA (right).
    Dark green indicates large gains; orange or yellow near the color-scale
    midpoint indicates conditions near zero or negative gain, concentrated in
    the brightness, fog, and contrast rows at low severity.}
  \Description{Three heatmaps showing accuracy differences from the source
    model by corruption type and severity for BN-adapt, TENT, and EATA.}
  \label{fig:condition-heatmaps}
\end{figure*}

Averages conceal individual conditions where adaptation hurts.
Figure~\ref{fig:condition-heatmaps} shows, for each method, the per-corruption,
per-severity accuracy delta against source as a heatmap; the three panels are
visually near-identical, indicating that the pattern of where adaptation helps
or harms is driven primarily by the corruption condition itself rather than by
which of the three methods is used. Table~\ref{tab:harm-rate} quantifies this:
6 to 7 of the 75 conditions show negative deltas for each method. These harmed
conditions are concentrated in a small set of low-severity, appearance-based
corruptions---brightness (severities 1--3), fog (severity 1, and severity 2 for
BN-adapt/EATA), contrast (severity 1), and defocus blur (severity 1)---where the
source model is already at or above 90\% accuracy and there is essentially no
distribution shift for the adaptation objective to correct.

\begin{table}[t]
  \caption{Share of corruption--severity conditions where each method's
    accuracy falls below the unadapted source model.}
  \label{tab:harm-rate}
  \centering
  \begin{tabular}{lcc}
    \toprule
    Method & Harm rate & Conditions harmed \\
    \midrule
    BN-adapt & 9.3\% & 7/75 \\
    TENT & 8.0\% & 6/75 \\
    EATA & 9.3\% & 7/75 \\
    \bottomrule
  \end{tabular}
\end{table}

Within these harmed conditions, TENT's harm is consistently the largest in
magnitude of the three methods (e.g., $-1.14$ pp at brightness severity 1,
versus $-0.79$ pp for BN-adapt and $-0.89$ pp for EATA), which is consistent
with TENT performing an additional gradient-based update on top of the same
statistic re-estimation that BN-adapt performs, giving entropy minimization
more room to act on a batch that did not need correcting.

\subsection{Worst-Case Robustness}

Beyond the average case, we examine the single hardest condition in the
benchmark: impulse noise at severity 5, where the source model achieves only
27.08\% accuracy. All three adaptation methods more than double this, with TENT
reaching 66.96\% (a 39.88-point recovery), compared with 63.73\% for BN-adapt
(+36.65 pp) and 63.96\% for EATA (+36.88 pp). TENT's largest margin over the
other two methods in the entire study occurs on precisely this worst-case
condition, suggesting that whatever benefit its additional gradient-based
update provides over simple statistic re-estimation is most pronounced when
the source model is failing badly---the mirror image of the low-severity
failure conditions, where the source model is already correct.

\begin{table}[t]
  \caption{Accuracy on the single hardest corruption--severity condition in
    the benchmark (impulse noise, severity 5).}
  \label{tab:worst-case}
  \centering
  \begin{tabular}{lc}
    \toprule
    Model or method & Accuracy (\%) \\
    \midrule
    Source & 27.08 \\
    BN-adapt & 63.73 \\
    TENT & \textbf{66.96} \\
    EATA & 63.96 \\
    \bottomrule
  \end{tabular}
\end{table}

\subsection{Batch-Size Sensitivity}

\begin{figure}[t]
  \centering
  \includegraphics[width=\linewidth]{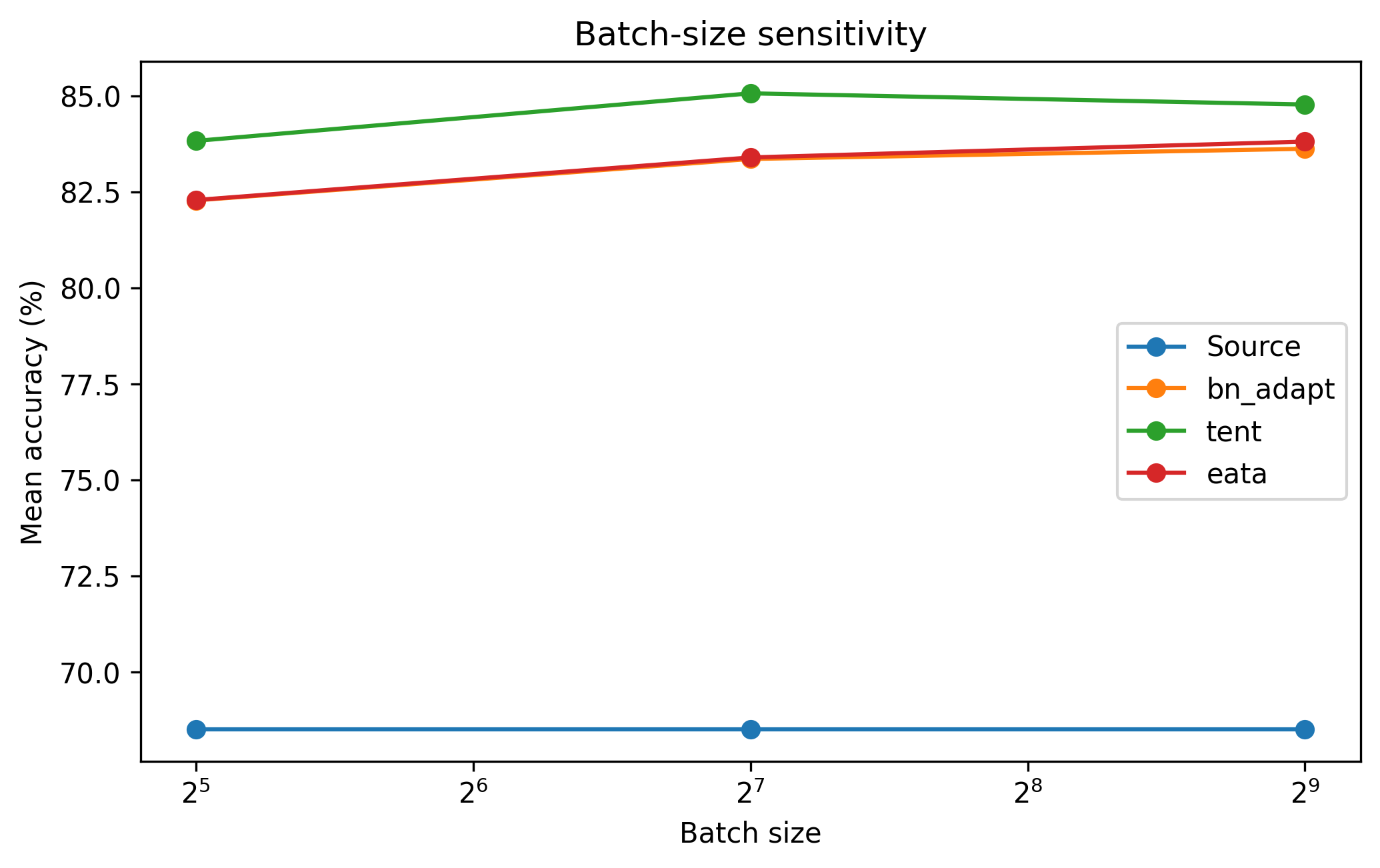}
  \caption{Mean accuracy versus test batch size (base-2 logarithmic scale).
    BN-adapt and EATA rise monotonically; TENT peaks at batch size 128 and dips
    slightly at 512.}
  \Description{Line chart of mean accuracy against batch sizes 32, 128, and 512
    for the source model and three adaptation methods.}
  \label{fig:batch-size}
\end{figure}

Figure~\ref{fig:batch-size} and Table~\ref{tab:batch-size} show mean accuracy at
test batch sizes of 32, 128, and 512, averaged over the six representative
corruptions at severities 3 and 5. BN-adapt and EATA both improve monotonically
with batch size in all 12 individual corruption--severity combinations tested,
consistent with larger batches giving a less noisy estimate of
batch-normalization statistics. TENT improves monotonically in only 5 of the 12
combinations: on the remaining 7, accuracy peaks at batch size 128 and then
falls slightly at 512 (for example, contrast at severity 5: 88.30\% at batch
32, 88.38\% at batch 128, and 87.79\% at batch 512).

Because larger batches also mean fewer gradient updates for a fixed number of
test images, this is consistent with TENT's entropy-minimization update trading
off statistic quality against the number of adaptation steps taken, a trade-off
that does not affect the statistics-only BN-adapt baseline and is strongly
attenuated for the filtered EATA variant in this reproduction. We report this
as an observed pattern in the current data rather than a confirmed mechanism,
since isolating the cause would require directly varying the number of gradient
steps independently of batch size.

\begin{table}[t]
  \caption{Mean accuracy by test batch size, averaged across six corruptions
    (Gaussian noise, motion blur, fog, contrast, JPEG compression, and snow) at
    severities 3 and 5.}
  \label{tab:batch-size}
  \centering
  \small
  \begin{tabular}{lrrrr}
    \toprule
    Batch size & Source & BN-adapt & TENT & EATA \\
    \midrule
    32 & 68.51 & 82.27 & \textbf{83.82} & 82.28 \\
    128 & 68.51 & 83.35 & \textbf{85.06} & 83.39 \\
    512 & 68.51 & 83.62 & \textbf{84.77} & 83.80 \\
    \bottomrule
  \end{tabular}
\end{table}

\subsection{Stability Under Continual, Long-Stream Adaptation}

A natural concern with any method that updates parameters online without
resetting to the source model is that errors could accumulate over a long
deployment, causing accuracy to collapse.

\begin{figure*}[t]
  \centering
  \includegraphics[width=0.86\textwidth]{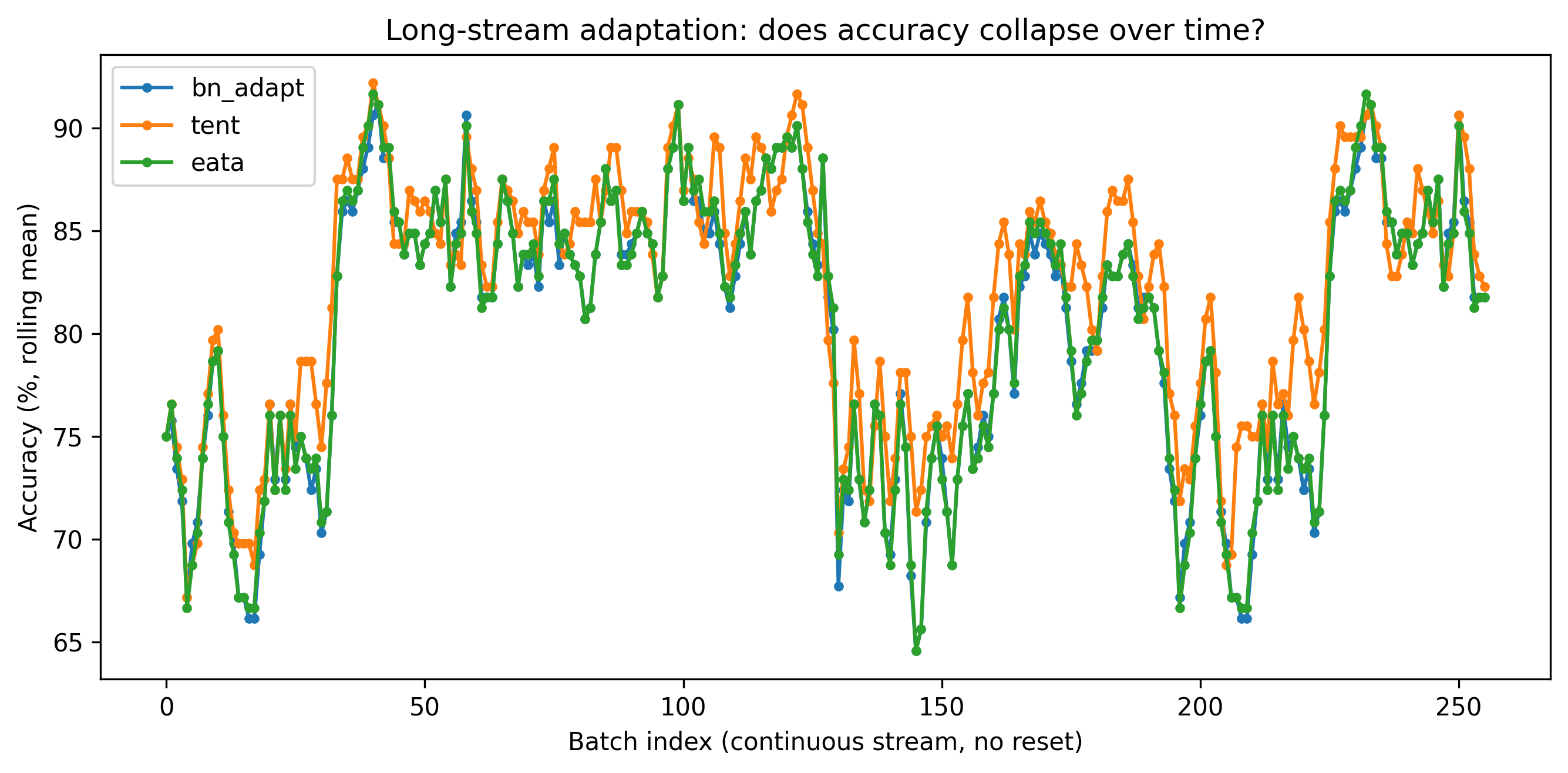}
  \caption{Rolling-mean accuracy across the 256-batch continual adaptation
    stream with no state reset. Sharp swings track corruption-block boundaries
    rather than a sustained drift; no method shows a persistent downward
    trend.}
  \Description{Line chart of rolling accuracy across 256 batches for BN-adapt,
    TENT, and EATA.}
  \label{fig:long-stream}
\end{figure*}

We test this directly using a 256-batch stream that cycles through six
corruptions at severity 5 without resetting the adapted model. Gaussian noise
and motion blur each occur twice, with their second occurrences separated from
the first by 160 intervening batches of adaptation on other corruptions.
Figure~\ref{fig:long-stream} shows the rolling-mean accuracy across the full
stream; none of the three methods exhibits a sustained downward drift.

Isolating the two recurring corruptions provides a cleaner comparison than a
simple first-half-versus-second-half analysis, since the corruption composition
differs across the two halves of the stream. As shown in
Table~\ref{tab:long-stream}, BN-adapt and EATA achieve effectively identical
accuracy on the first and second occurrences of each repeated corruption, with
differences within 0.1~pp. TENT, which carries its adapted affine parameters
across the intervening batches, remains stable or improves modestly on the
second occurrence, from 74.37\% to 76.37\% on Gaussian noise and from 86.52\%
to 86.67\% on motion blur. We find no evidence of accuracy collapse for any
method over the 256-batch stream tested.

\begin{table*}[t]
  \caption{Mean batch accuracy on the first versus second occurrence of the
    same corruption within the 256-batch continual stream, separated by 160
    intervening batches of adaptation on other corruptions.}
  \label{tab:long-stream}
  \centering
  \begin{tabular}{llrrr}
    \toprule
    Corruption & Method & First occurrence (\%) & Second occurrence (\%) & $\Delta$ (pp) \\
    \midrule
    Gaussian noise & BN-adapt & 72.46 & 72.46 & 0.00 \\
    Gaussian noise & TENT & 74.37 & 76.37 & +2.00 \\
    Gaussian noise & EATA & 72.46 & 72.46 & 0.00 \\
    Motion blur & BN-adapt & 85.94 & 85.94 & 0.00 \\
    Motion blur & TENT & 86.52 & 86.67 & +0.15 \\
    Motion blur & EATA & 86.04 & 86.04 & 0.00 \\
    \bottomrule
  \end{tabular}
\end{table*}

\section{Discussion}

Taken together, these observations support a specific and fairly narrow claim:
entropy minimization, as implemented in TENT, delivers the largest average
benefit of the three methods studied here, and its advantage is most pronounced
exactly where it is needed most---on the noisiest, highest-severity, worst-case
conditions. The same mechanism that produces this advantage, however, also
produces the study's clearest failure pattern: on a small number of
low-severity, appearance-only corruptions where the source model is already
accurate, TENT's additional gradient-based update on top of statistic
re-estimation causes slightly more harm than BN-adapt's statistics-only
approach.

A plausible reading of this pattern is that entropy minimization implicitly
assumes the current prediction distribution is one worth sharpening; when the
source model's predictions are already close to correct and confident,
sharpening them further has little room to help and some room to overshoot,
whereas the batch-normalization re-estimation shared by all three methods is
closer to a no-op in that regime and correspondingly safer.

The batch-size result points to a related but distinct issue: TENT is the only
one of the three methods whose accuracy is not a monotonic function of batch
size, which matters practically because test batch size is often set by memory
or latency constraints rather than chosen to maximize accuracy. Since BN-adapt
and EATA both improve monotonically under the same conditions, this
non-monotonicity appears specific to TENT's gradient-based update rather than
to batch-normalization statistic quality in general. The long-stream result,
by contrast, is reassuring: none of the three methods shows accuracy collapse
over 256 batches of continuous, unreset adaptation, which is consistent with
EATA's design intent and suggests TENT is not accumulating harmful drift
either, at least at this stream length and under this corruption schedule.

None of these observations point to a specific alternative algorithm, and we
do not propose one. What they do suggest, descriptively, is that a fixed and
uniformly applied entropy-minimization signal treats all batches as equally
worth adapting to, when the evidence here indicates that batches differ
systematically---by corruption type, by severity, and by how many samples
happen to be in them---in how much they benefit from or are harmed by that
signal. Whether some estimate of per-batch or per-sample reliability could be
used to modulate the strength of the entropy-minimization update, and whether
doing so would meaningfully change the failure conditions identified here
without eroding the worst-case gains, is a natural question raised by these
data but not one we answer here.

\section{Limitations}

\begin{itemize}
  \item All results are on a single benchmark (CIFAR-10-C) and a single source
    model. We do not report CIFAR-100-C or ImageNet-C results, and
    generalization of the specific failure conditions to other architectures
    or datasets is untested.
  \item The batch-size and long-stream protocols were run on a representative
    subset of six corruptions rather than the full set of 15, chosen for
    computational cost. The batch-size monotonicity result in particular should
    be checked against the remaining corruption types before being treated as
    general.
  \item The statistical tests compare paired condition-level mean accuracies
    ($n=75$), not individual samples. Per-sample entropy, margin, or confidence
    logs were not collected, which limits our ability to characterize why
    individual predictions change.
  \item CoTTA and SAR are discussed as related work
    (Section~\ref{subsec:continual-related-work}) but neither was included in
    the quantitative comparison; the harm-rate, batch-size, and long-stream
    findings reported here are specific to TENT, BN-adapt, and EATA as
    implemented in this reproduction.
  \item The long-stream protocol covers 256 batches at severity 5. Longer
    streams, mixed-severity streams, and abrupt rather than blocked corruption
    switching were not tested.
\end{itemize}

\section{Conclusion}

We reproduced TENT, BN-adapt, and EATA on CIFAR-10-C under a common protocol
and examined their behavior beyond the single aggregate accuracy number
typically reported. The reproduction confirms the central claim in the TTA
literature: entropy minimization improves accuracy over an unadapted source
model, with the benefit growing sharply as corruption severity increases.
Looking beneath that average, however, we find a consistent, if small, set of
conditions---low-severity appearance corruptions---where every method studied,
and TENT most of all, performs worse than doing nothing; a batch-size dependence
specific to TENT's gradient-based update; and no evidence of long-stream
collapse for any method at the stream length tested. We present these as
documented empirical observations rather than as a new method, in the hope
that they usefully narrow the space of conditions that a future
reliability-aware or condition-aware adaptation strategy would need to address.


\begin{thebibliography}{1}

\bibitem{hendrycks2019}
Dan Hendrycks and Thomas Dietterich. 2019.
\newblock Benchmarking neural network robustness to common corruptions and
perturbations.
\newblock In \emph{International Conference on Learning Representations}.

\bibitem{krizhevsky2009}
Alex Krizhevsky. 2009.
\newblock \emph{Learning Multiple Layers of Features from Tiny Images}.
\newblock Technical Report, University of Toronto.

\bibitem{nado2020}
Zachary Nado, Shreyas Padhy, D. Sculley, David D'Amour, Balaji
Lakshminarayanan, and Jasper Snoek. 2020.
\newblock Evaluating prediction-time batch normalization for robustness under
covariate shift.
\newblock arXiv:2006.10963.

\bibitem{niu2022eata}
Shuaicheng Niu, Jiaxiang Wu, Yifan Zhang, Yaofo Chen, Shijian Zheng, Peilin
Zhao, and Mingkui Tan. 2022.
\newblock Efficient test-time model adaptation without forgetting.
\newblock In \emph{International Conference on Machine Learning}.

\bibitem{schneider2020}
Steffen Schneider, Evgenia Rusak, Luisa Eck, Oliver Bringmann, Wieland Brendel,
and Matthias Bethge. 2020.
\newblock Improving robustness against common corruptions by covariate shift
adaptation.
\newblock In \emph{Advances in Neural Information Processing Systems}.

\bibitem{wang2021tent}
Dequan Wang, Evan Shelhamer, Shaoteng Liu, Bruno Olshausen, and Trevor Darrell.
2021.
\newblock Tent: Fully test-time adaptation by entropy minimization.
\newblock In \emph{International Conference on Learning Representations}.

\bibitem{wang2022cotta}
Qin Wang, Olga Fink, Luc Van Gool, and Dengxin Dai. 2022.
\newblock Continual test-time domain adaptation.
\newblock In \emph{IEEE/CVF Conference on Computer Vision and Pattern
Recognition}.

\bibitem{zagoruyko2016}
Sergey Zagoruyko and Nikos Komodakis. 2016.
\newblock Wide residual networks.
\newblock In \emph{British Machine Vision Conference}.

\bibitem{iwasawa2021t3a}
Yusuke Iwasawa and Yutaka Matsuo. 2021.
\newblock Test-time classifier adjustment module for model-agnostic domain
generalization.
\newblock In \emph{Advances in Neural Information Processing Systems}.

\bibitem{liang2020shot}
Jian Liang, Dapeng Hu, and Jiashi Feng. 2020.
\newblock Do we really need to access the source data? Source hypothesis transfer for unsupervised domain adaptation.
\newblock In \emph{International Conference on Machine Learning}, pages
6028--6039.

\bibitem{niu2023sar}
Shuaicheng Niu, Jiaxiang Wu, Yifan Zhang, Zhiquan Wen, Yaofo Chen, Peilin Zhao,
and Mingkui Tan. 2023.
\newblock Towards stable test-time adaptation in dynamic wild world.
\newblock In \emph{International Conference on Learning Representations}.

\bibitem{wang2023otta}
Zixin Wang, Yadan Luo, Liang Zheng, Zhuoxiao Chen, Sen Wang, and Zi Huang. 2024.
\newblock In search of lost online test-time adaptation: A survey.
\newblock \emph{arXiv preprint arXiv:2310.20199}.

\bibitem{xiao2024ttasurvey}
Zehao Xiao and Cees G. M. Snoek. 2024.
\newblock Beyond model adaptation at test time: A survey.
\newblock \emph{arXiv preprint arXiv:2411.03687}.

\bibitem{croce2021robustbench}
Francesco Croce, Maksym Andriushchenko, Vikash Sehwag, Edoardo Debenedetti,
Nicolas Flammarion, Mung Chiang, Prateek Mittal, and Matthias Hein. 2021.
\newblock RobustBench: A standardized adversarial robustness benchmark.
\newblock In \emph{Advances in Neural Information Processing Systems Datasets
and Benchmarks Track}.

\bibitem{kingma2014adam}
Diederik P. Kingma and Jimmy Ba. 2015.
\newblock Adam: A method for stochastic optimization.
\newblock In \emph{International Conference on Learning Representations}.

\end{thebibliography}

\end{document}